\documentclass[runningheads]{llncs}
\usepackage[outdir=./]{epstopdf}
\usepackage{graphicx}

\usepackage{indentfirst}

\begin{document}
\title{BERT-Based Multi-Head Selection for Joint Entity-Relation Extraction}
%
%
\author{Weipeng Huang, Xingyi Cheng, Taifeng Wang, Wei Chu}
\institute{Ant Financial Services Group
\email{\{weipeng.hwp,fanyin.cxy,taifeng.wang,weichu.cw\}@antfin.com}}

%
%

\maketitle              
\begin{abstract}

In this paper, we report our method for the Information Extraction task in 2019 Language and Intelligence Challenge. We incorporate BERT into the multi-head selection framework for joint entity-relation extraction. This model extends existing approaches from three perspectives. First, BERT is adopted as a feature extraction layer at the bottom of the multi-head selection framework. We further optimize BERT by introducing a semantic-enhanced task during BERT pre-training. Second, we introduce a large-scale Baidu Baike corpus for entity recognition pre-training, which is of weekly supervised learning since there is no actual named entity label. Third, soft label embedding is proposed to effectively transmit information between entity recognition and relation extraction. Combining these three contributions, we enhance the information extracting ability of the multi-head selection model and achieve F1-score 0.876 on testset-1 with a single model. By ensembling four variants of our model, we finally achieve F1 score 0.892 (1st place) on testset-1 and F1 score 0.8924 (2nd place)  on testset-2.

\end{abstract}

\section{Problem Definition}
Given a sentence and a list of pre-defined schemas which define the relation P and the classes of its corresponding subject S and object O, for example, (S\_TYPE: Person, P: wife, O\_TYPE: Person), (S\_TYPE: Company, P: founder, O\_TYPE: Person), a participating information extraction (IE) system is expected to output all correct triples [(S1, P1, O1), (S2, P2, O2) ...] mentioned in the sentence under the constraints of given schemas. A largest schema-based Chinese information extraction dataset is released in this competition. Precision, Recall and F1 score are used as the basic evaluation metrics to measure the performance of participating systems. 

	\begin{figure}
	\centering
	\vspace*{-2mm}
	\includegraphics[width=0.9\textwidth]{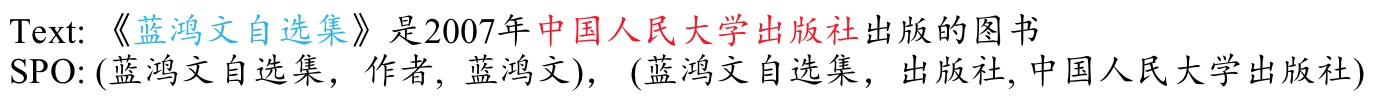}
	\vspace*{-2mm}
	\caption{An example in the dataset.}
	\label{example}
\end{figure}

From the example shown in Figure~\ref{example}, we can notice that one entity can be involved in multiple triplets and entity spans have overlaps, which is the difficulties of this task.

\section{Related Work}
Recent years, great efforts have been made on extracting relational fact from unstructured raw texts to build large structural knowledge bases. A relational fact is often represented as a triplet which consists of two entities (subject and object) and semantic relation between them. Early works \cite{ref_Semeval,ref_Zeng,ref_Xu} mainly focused on the task of relation classification which assumes the entity pair are identified beforehand. This limits their practical application since they neglect the extraction of entities. To extract both entities and their relation, existing methods can be divided into two categories : the pipelined framework, which first uses sequence labeling models to extract entities, and then uses relation classification models to identify the relation between each entity pair; and the joint approach, which combines the entity model and the relation model through different strategies, such as constraints or parameters sharing.

\vspace*{2mm} 
\subsubsection{Pipelined framework} 
\begin{figure}
	\centering
	\vspace*{-2mm}
	\includegraphics[width=0.93\textwidth]{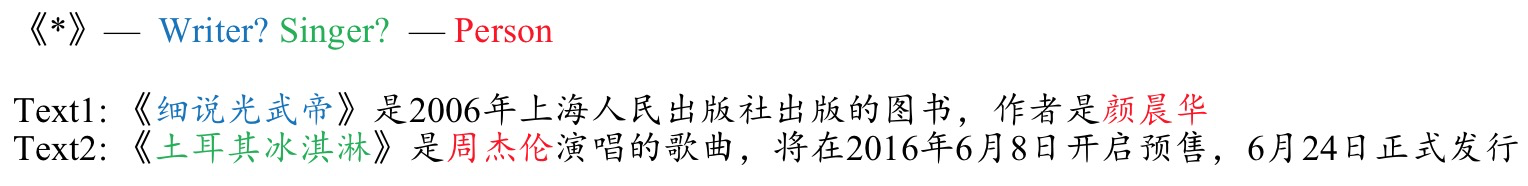}
	\vspace*{-2mm}
	\caption{Examples: entity contained in book title marks can be song or book, its relation to a person can be singer or writer.}
	\label{two_example}
\end{figure}

Many earlier entity-relation extraction systems \cite{ref_Zelenko,ref_MiwaMakoto,ref_Chan} adopt pipelined framework: they first conduct entity extraction and then predict the relations between each entity pair. The pipelined framework has the flexibility of integrating different data sources and learning algorithms, but their disadvantages are obvious. First, they suffer significantly from error propagation, the error of the entity extraction stage will be propagated to the relation classification stage.  Second, they ignore the relevance of entity extraction and relation classification. As shown in Figure ~\ref{two_example}, entity contained in book title marks can be a song or book, its relation to a person can be singer or writer. Once the relationship has been confirmed, the entity type can be easily identified, and vice versa. For example, if we know the relationship is singer, then the entity type should be a song. Entity extraction and relation classification can benefit from each other so it will harm the performance if we consider them separately. Third, the pipelined framework results in low computational efficiency. After the entity extraction stage, each entity pair should be passed to the relation classification model to identify their relation. Since most entity pairs have no relation, this two-stage manner is inefficient.

\vspace*{2mm} 
\subsubsection{Joint model} To overcome the aforementioned disadvantages of the pipelined framework, joint learning models have been proposed. Early works \cite{ref_Yu,ref_Miwa,ref_Li} need a complicated process of feature engineering and heavily depends on NLP tools for feature extraction. Yu and Lam (2010) \cite{ref_Yu}  proposed the approach to connect the two models through global probabilistic graphical models. Li and Ji (2014) \cite{ref_Li} extract entity mentions and relations using structured perceptron with efficient beam search, which is significantly more efficient and less time-consuming than constraint-based approaches. Gupta et al. (2016) \cite{ref_Gupta} proposed the table-filling approach, which provides an opportunity to incorporate more sophisticated features and algorithms into the model, such as search orders in decoding and global features.

 Neural network models have been widely used in the literature as well.  Zheng  et al. (2017) \cite{ref_Zheng} propose a novel tagging scheme that can convert the joint extraction task to a tagging problem. This tagging based method is better than most of the existing pipelined methods, but its flexibility is limited and can not tackle the situations when (1) one entity belongs to multiple triplets (2) multiple entities have overlaps.  Zeng  et al. (2018) \cite{ref_Xiangrong}  propose an end2end neural model based on sequence-to-sequence learning with copy mechanism to extract relational facts from sentences, where the entities and relations could be jointly extracted. The performance of this method is limited by the word segmentation accuracy because it can not extract entities beyond the word segmentation results. Li et al. \cite{ref_Xiaoya} (2019) cast the task as a multi-turn question answering problem, i.e., the extraction of entities and relations is transformed to the task of identifying answer spans from the context. This framework provides an elegant way to capture the hierarchical dependency of tags. However, it is also of low computational efficiency since it needs to scan all entity template questions and corresponding relation template questions for a single sentence.  Bekoulis et al. (2017) \cite{ref_Bekoulis} propose a joint neural model which performs entity recognition and relation extraction simultaneously, without the need of any manually extracted features or the use of any external tool. They model the entity recognition task using a CRF (Conditional Random Fields) layer and the relation extraction task as a multi-head selection
 problem since one entity can have multiple relations. The model adopted BiLSTM to extract contextual feature and propose a label embedding layer to connect the entity recognition branch and the relation classification branch. Our model is based on this framework and make three improvements: 
 
 (1) BERT \cite{ref_bert} is introduced as a feature extraction layer in place of BiLSTM. We also optimize the pre-training process of BERT by introducing a semantic-enhanced task.
 
 (2) A large-scale Baidu Baike corpus is introduced for entity recognition pre-training, which is of weekly supervised learning since there is no actual named entity label.
 
 (3) Soft label embedding is proposed to effectively transmit information between entity recognition and relation extraction.

%
%
%
%

\section{Model Description}

\subsection{Overall Framwork}

\begin{figure}
	\centering
	\vspace*{-2mm}
	\includegraphics[width=1.0\textwidth]{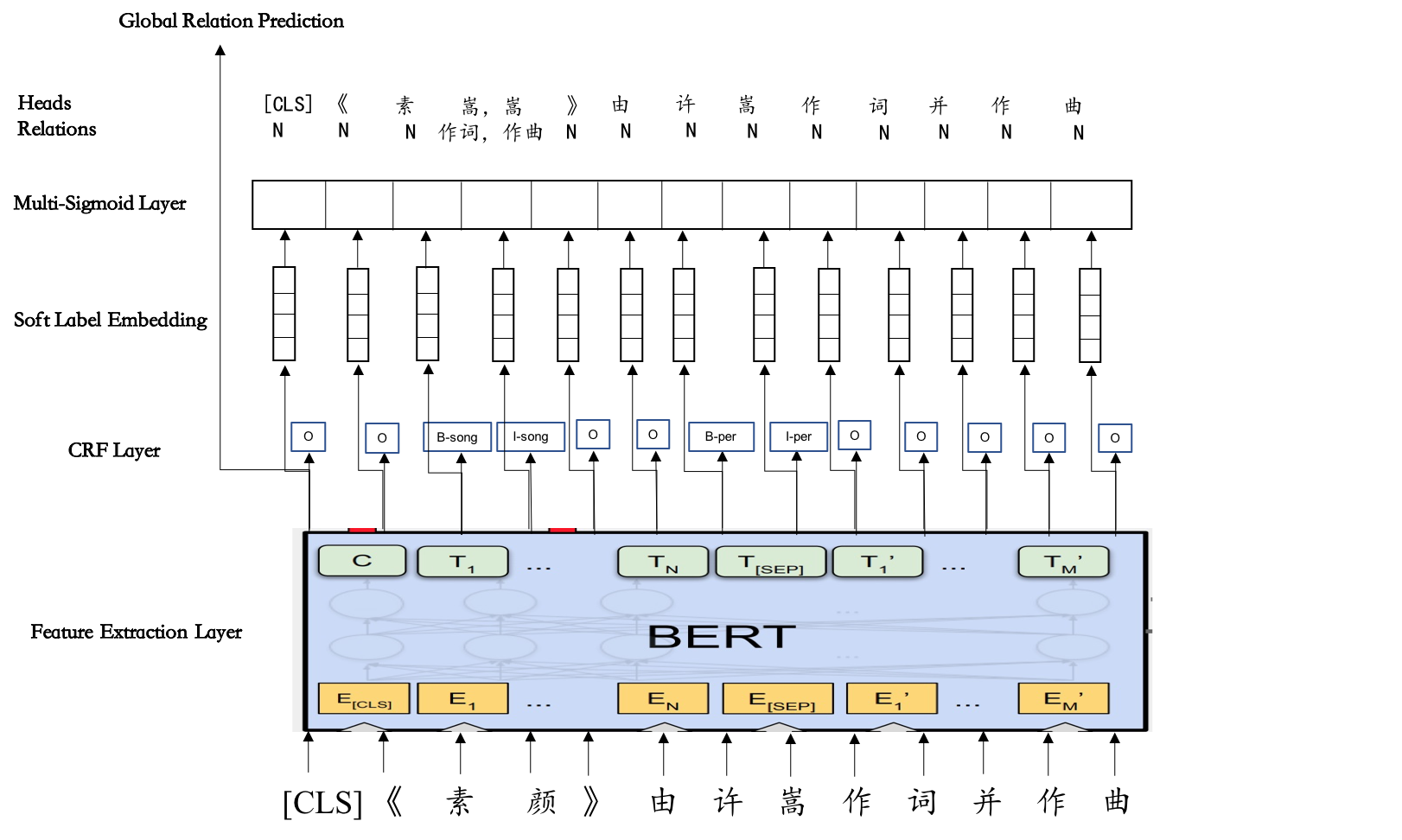}
	\vspace*{-2mm}
	\caption{Overall Framwork: BERT-Based Multi-Head Selection}
	\label{framework}
\end{figure}

Figure ~\ref{framework}  summarizes the proposed model architecture. The model takes character sequence as input and captures contextual features using BERT.  A CRF layer is applied to extract entities from the sentence. To effectively transmit information between entity recognition and relation extraction, soft label embedding is built on the top of CRF logits. To solve the problem that one entity belongs to multiple triplets, a multi-sigmoid layer is applied. We find that adding an auxiliary global relation prediction task also improve the performance.

\subsection{BERT for Feature Extraction}
BERT (Bidirectional Encoder Representations from Transformers) \cite{ref_bert} is a new language representation model, which uses bidirectional transformers to pre-train a large unlabeled corpus, and fine-tunes
the pre-trained model on other tasks. BERT has been widely used and shows great improvement on various natural language processing tasks, e.g., word segmentation, named entity recognition, sentiment analysis, and question answering. We use BERT to extract contextual feature for each character instead of BiLSTM in the original work \cite{ref_Bekoulis}. To further improve the performance, we optimize the pre-training process of BERT by introducing a semantic-enhanced task.

\subsubsection{Enhanced BERT} Original google BERT is pre-trained using two unsupervised tasks, masked language model (MLM) and next sentence prediction (NSP). MLM task enables the model to capture the discriminative contextual feature. NSP task makes it possible to understand the relationship between sentence pairs, which is not directly captured by language modeling. We further design a semantic-enhanced task to enhance the performance of BERT. It incorporate previous sentence prediction and document level prediction. We pre-train BERT by combining MLM, NSP and the semantic-enhanced task together \cite{ref_symbert}.

\subsection{Named Entity Recognition}

NER (Named Entity Recognition) is the first task in the joint multi-head selection model. It is usually formulated as a sequence labeling problem using the BIO (Beginning, Inside, Outside) encoding scheme. Since there are different entity types, the tags are extended to B-type, I-type and O. Linear-chain CRF \cite{ref_Lafferty} is widely used for sequence labeling in deep models. In our method, CRF is built on the top of BERT. Supposed $y\in {\left \{ B-type,I-type,O \right \}}$ is the label, score function $ s(X,i)_{y_{i}} $ is the output of BERT at $ i_{th}$ character and $ b_{y_{i-1}y_{i}} $ is trainable parameters, the probability of a possible label sequence is formalized as:

\begin{equation}
P(Y|X)=\frac{\prod_{i=2}^{n}exp(s(X,i)_{y_{i}}+b_{y_{i-1}y_{i}})))}{\sum_{y^{'}}\prod_{i=2}^{n}exp(s(X,i)_{y_{i}^{'}}+b_{y_{i-1}^{'}y_{i}^{'}})))}
\end{equation}

By solving Eq~\ref{eq-2} we can obtain the optimal sequence tags:

\begin{equation}
\label{eq-2}
Y^{*} = argmax P(Y|X)
\end{equation}


\begin{figure}
	\centering
	\vspace*{-2mm}
	\includegraphics[width=0.9\textwidth]{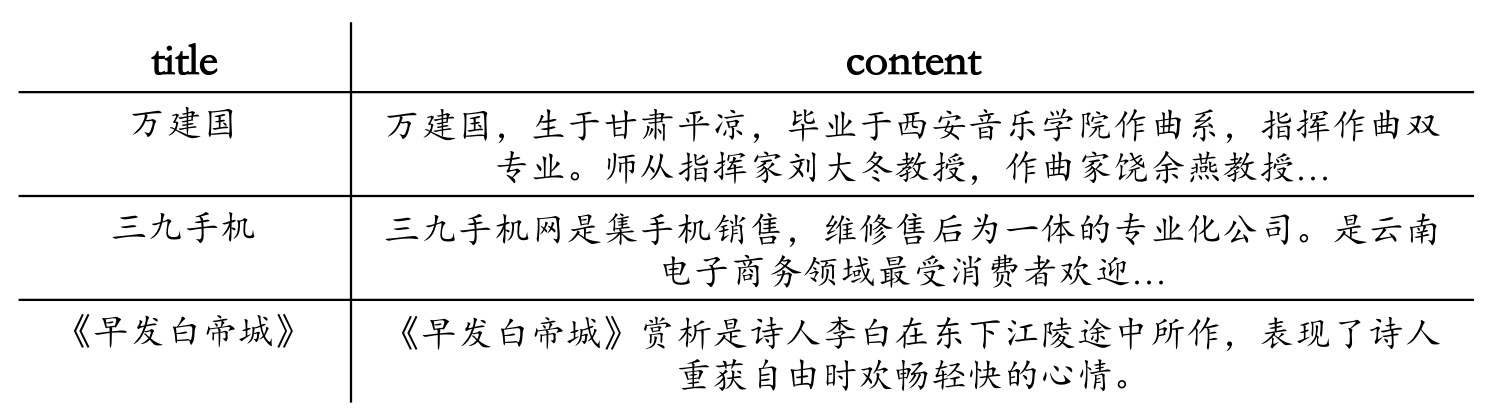}
	\vspace*{-2mm}
	\caption{Crawled corpus from Baidu Baike.}
	\label{crawl}
\end{figure}

\subsubsection{Extra Corpus for NER Pretraining}
Previous works show that introducing extra data for distant supervised learning usually boost the model performance. For this task, we collect a large-scale Baidu Baike corpus (about 6 million sentences) for NER pre-training. As shown in figure ~\ref{crawl}, each sample contains the content and its title. These samples are auto-crawled so there is no actual entity label. We consider the title of each sample as a pseudo label and conduct NER pre-training using these data. Experimental results show that it improves performance.

\subsection{Soft Label Embedding}
Miwa et al.  (2016)  \cite{ref_Makoto} and Bekoulis et al. (2018) \cite{ref_Bekoulis} use the entity tags as input to relation classification layer by learning label embeddings. As reported in their experiments, an improvement of 1$\sim$2\% F1 is achieved with the use of label embeddings. Their mechanism is hard label embedding because they use the CRF decoding results, which have two disadvantages. On one hand, the entity recognition results are not absolutely correct since they are predicted by the model during inference. The error from the entity tags may propagate to the relation classification branch and hurt the performance. On the other hand, CRF decoding process is based on the Viterbi Algorithm, which contains an argmax operation which is not differentiable. To solve this problem, we proposed soft label embedding, which takes the logits as input to preserve probability of each entity type. Suppose $N$ is the  logits dimension, i.e., the number of entity type, \textbf{M} is the label embedding matrix, then soft label embedding for $ i_{th}$ character can be formalized as Eq~\ref{eq-3}:

\begin{equation}
\label{eq-3}
h_i = \frac{\sum softmax(s(X,i))\cdot\textbf{M}}{N}
\end{equation}

\subsection{Relation Classification as Multi-Head Selection}
We formulated the relation classification task as a multi-head selection problem, since each token in the sentence has multiple heads, i.e., multiple relations  with other tokens. Soft label embedding of the $ i_{th}$ token $ h_{i}$ is feed into two separate fully connected layers to get the subject representation  $ h_{i}^{s}$ and object representation $ h_{i}^{o}$. Given the $ i_{th}$ token ($ h_{i}^{s}$, $ h_{i}^{o}$) and the $ j_{th}$ token ($ h_{j}^{s}$, $ h_{j}^{o}$) , our task is to predict their relation: 

\begin{equation}
\label{eq-4}
 r_{i,j} = f(h_{i}^{s},h_{j}^{o}) ,  r_{j,i} = f(h_{j}^{s},h_{i}^{o}) 
\end{equation}

where $f(\cdot)$  means neural network, $ r_{i,j}$ is the relation when  the $ i_{th}$ token is subject and the $ j_{th}$ token is object, $ r_{j,i}$ is the relation when  the $ j_{th}$ token is subject and the $ i_{th}$ token is object. Since the same entity pair have multiple relations, we adopt multi-sigmoid layer for the relation prediction. We minimize the cross-entropy loss $L_{rel}$ during training:

\begin{equation}
\label{eq-5}
L_{rel} = \sum_{i=0}^{K} \sum_{j=0}^{K} NLL(r_{i,j}, y_{i,j})
\end{equation}

where $K$ is the sequence length and $y_{i,j}$ is ground truth relation label.
 
\subsubsection{Global Relation Prediction}
Relation classification is of entity pairs level in the original multi-head selection framework. We introduce an auxiliary sentence-level relation classification prediction task to guide the feature learning process. As shown in figure  ~\ref{framework}, the final hidden state of the first token $[CLS]$ is taken to obtain a fixed-dimensional pooled representation of the input sequence. The hidden state is then feed into a multi-sigmoid layer for classification. In conclusion, our model is trained using the combined loss:

\begin{equation}
\label{eq-6}
L = L_{ner} + L_{rel} + L_{global\_rel}
\end{equation}

\subsection{Model Ensemble}
Ensemble learning is an effective method to further improve performance. It is widely used in data mining and machine learning competitions. The basic idea is to combine the decisions from multiple models to improve the overall performance. In this work, we combine four variant multi-head selection models by learning an XGBoost \cite{ref_Tianqi} binary classification model on the development set. Each triplet generated by the base model is treated as a sample. We then carefully design 200-dimensional features for each sample. Take several important features for example:

$\cdot$ the probability distribution of the entity pair

$\cdot$ the probability distribution of sentence level

$\cdot$ whether the triplet appear in the training set

$\cdot$ the number of predicted entities, triples, relations of the given sentence

$\cdot$ whether the entity boundary is consistent with the word segmentation results 

$\cdot$ semantic feature. We contact the sentence and the triplet to train an NLI model, hard negative triplets are constructed to help NLI model capture semantic feature.

\section{Experiments}
\subsection{Experimental Settings}
All experiments are implemented on the hardware with Intel(R) Xeon(R) CPU E5-2682 v4 @ 2.50GHz and NVIDIA Tesla P100.

\subsubsection{Dataset and evaluation metrics}

We evaluate our method on the SKE dataset used in this competition, which is the largest schema-based Chinese information extraction dataset in the industry, containing more than 430,000 SPO triples in over 210,000 real-world Chinese sentences, bounded by a pre-specified schema with 50 types of predicates. All sentences in SKE Dataset are extracted from Baidu Baike and Baidu News Feeds. The dataset is divided into a training set (170k sentences), a development set (20k sentences) and a testing set (20k sentences). The training set and the development set are to be used for training and are available for free download. The test set is divided into two parts, the test set 1 is available for self-verification, the test set 2 is released one week before the end of the competition and used for the final evaluation.

\subsubsection{Hyperparameters}
The max sequence length is set to 128, the number of fully connected layer of relation classification branch is set to 2, and that of global relation branch is set to 1. During training, we use Adam with the learning rate of 2e-5, dropout probability of 0.1. This model converges in 3 epoch.

\subsubsection{Preprocessing}
All uppercase letters are converted to lowercase letters. We use max sequence length 128 so sentences longer than 128 are split by punctuation. According to FAQ, entities in book title mark should be completely extracted. Because the annotation criteria in trainset are diverse, we revise the incomplete entities. To keep consistence, book title marks around the entities are removed. 
 
\subsubsection{Postprocessing}
Our postprocessing mechanism is mainly based on the FAQ evaluation rules. After model prediction, we remove triplets whose entity-relation types are against the given schemas. For entities contained in book title mark, we complement them if they are incomplete. Date type entities are also complemented to the finest grain. These are implemented by regular expression matching.

Note that entity related preprocessing and postprocessing are also performed on the development set to keep consistency with the test set, thus the change of development metric is reliable.

\subsection{Main Results}
Results on SKE dataset are presented in Table 1. The baseline model is based on the Google BERT, use hard label embedding and train on only SKE dataset without NER pretraining. As shown in table 1, the F1 score increase from 0.864 to 0.871 when combined with our enhanced BERT. NER pretraining using the extra corpus, soft label embedding and auxiliary sentence-level relation classification prediction also improve the F1 score. Combined all of these contributions, we achieve F1-score 0.876 with the single model on test set 1.

\begin{table}
	\caption{Performance of variant multi-head selection model on SKE dataset.}\label{tab1}
	\small
	\centering

	\begin{tabular}{c|c|c|c|c|c|c}
		Model & dev-P & dev-R & dev-F1 & test1-P & test1-R & test1-F1\\
		\hline
		Baseline & 0.796 & 0.845 & 0.819 & 0.902 & 0.828 & 0.864   \\
		
		\hline
		Baseline+Enhanced BERT & 0.809& 0.854 & 0.830 & 0.872 & 0.870 & 0.871   \\
		
		\hline
		Baseline+NER Pretraining & 0.803 & 0.852 & 0.827 & 0.883 & 0.854 & 0.868   \\
		
		\hline
		Baseline+Soft label embedding & 0.814 & 0.832 & 0.823 & 0.868 & 0.866  & 0.867   \\
		
		\hline
		Baseline+Global Predicate Prediction& 0.806 & 0.838 & 0.822 & 0.891 & 0.842 & 0.866   \\
		
		\hline
		Baseline+all & 0.821 & 0.855 & 0.837 & 0.873 & 0.879 & 0.876   \\

	\end{tabular}
\end{table}

\begin{figure}
	\centering
	\vspace*{-2mm}
	\includegraphics[width=0.8\textwidth]{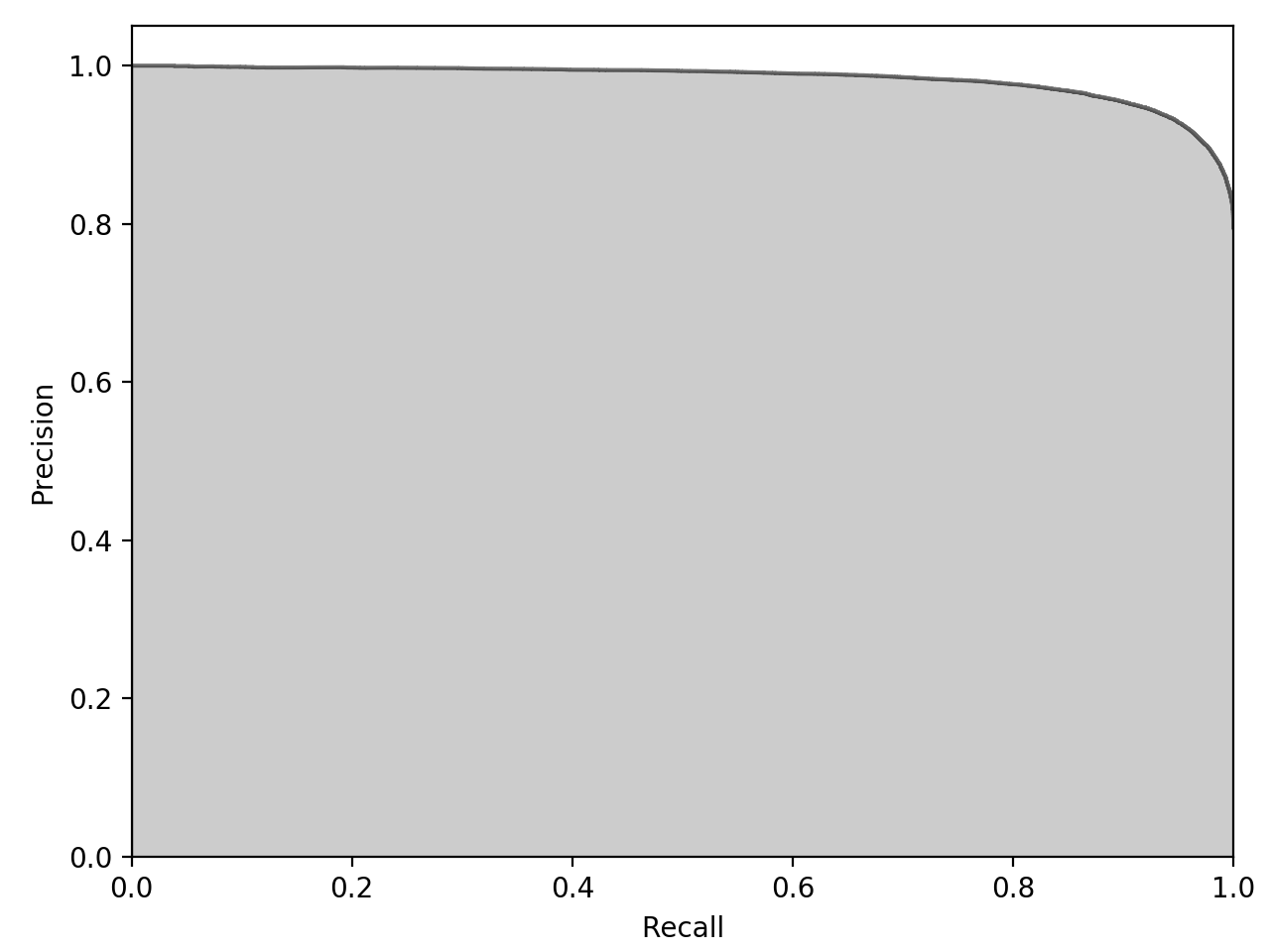}
	\vspace*{-2mm}
	\caption{Precision-Recall Curve}
	\label{pr}
\end{figure}

\subsection{Model Ensemble}
We select the following four variant model to further conduct model ensembling. The ensemble model is XGBoost binary classifier, which is very fast during training. Since the base models are trained on the training set, we perform cross-validation on development set,  figure ~\ref{pr} shows the PR curve of the ensemble model. By model ensembling the F1 score increase from 0.876 to 0.892.

$\cdot$ Google BERT + Soft Label Embedding + Global Relation Prediction

$\cdot$ Enhanced BERT + Soft Label Embedding + Global Relation Prediction

$\cdot$ Google BERT + Soft Label Embedding + Global Relation Prediction + NER Pretraining

$\cdot$ Enhance BERT + Soft Label Embedding + Global Relation Prediction  + NER Pretraining

\subsection{Case Study}
\begin{figure}
	\centering
	
	\includegraphics[width=1.0\textwidth]{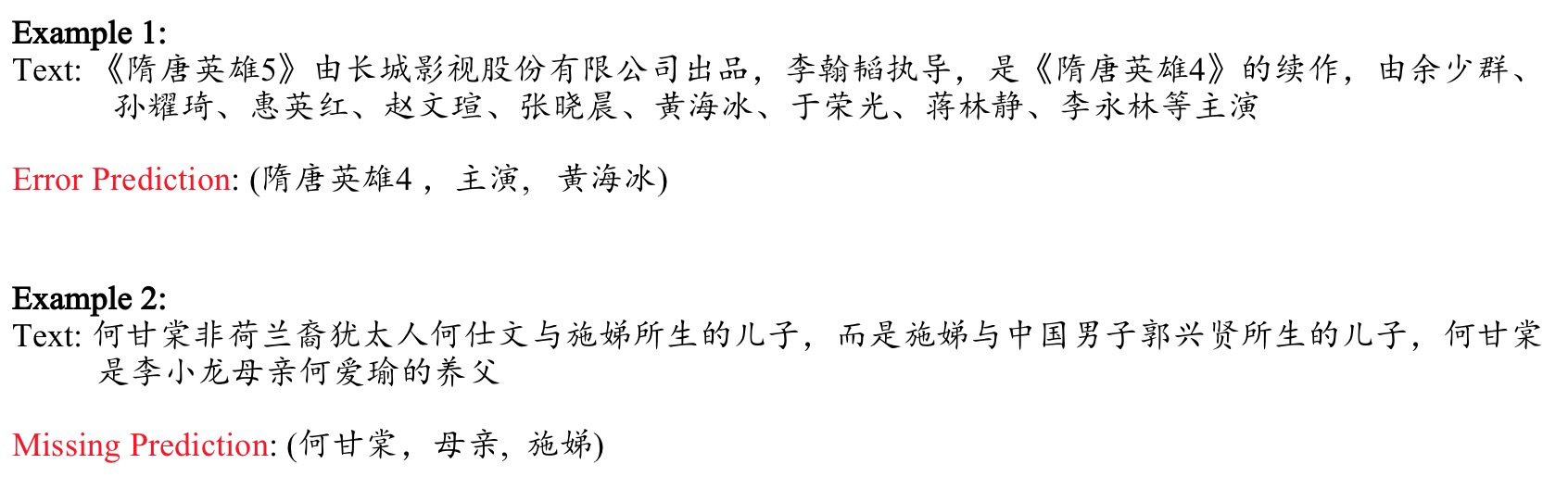}
	\caption{Examples the model fails to predict.}
	\label{case}
\end{figure}

Two examples of our model fail to predict are shown in figure ~\ref{case}. For example 1, the triplet can not be drawn from the given sentence. However, the triplet is actually in the trainset. Our model may overfit to the trainset in this situation. For example 2, there is complicate family relationships mentioned in the sentence, which is too hard for the model to capture. To solve this problem, a more robust model should be proposed and we leave this as future work.

\section{Conclusion}
In this paper, we report our solution to the information extraction task in 2019 Language and Intelligence Challenge. We first analyze the problem and find that most entities are involved in multiple triplets. To solve this problem, we incorporate BERT into the multi-head selection framework for joint entity-relation extraction. Enhanced BERT pre-training, soft label embedding and NER pre-training are three main technologies we introduce to further improve the performance. Experimental results show that our method achieves competitive performance: F1 score 0.892 (1st place) on the test set 1 and F1 score 0.8924 (2nd place)  on the test set 2.


\newpage
\begin{thebibliography}{8}

\bibitem{ref_bert}
Devlin, Jacob, et al. Bert: Pre-training of deep bidirectional transformers for language understanding.  arXiv preprint arXiv:1810.04805 (2018).

\bibitem{ref_Semeval}
Hendrickx, Iris, et al. "Semeval-2010 task 8: Multi-way classification of semantic relations between pairs of nominals." Proceedings of the Workshop on Semantic Evaluations: Recent Achievements and Future Directions. Association for Computational Linguistics, 2009.

\bibitem{ref_Zeng}
Zeng D, Liu K, Lai S, et al. Relation classification via convolutional deep neural network[J]. 2014.

\bibitem{ref_Xu}
Xu K, Feng Y, Huang S, et al. Semantic relation classification via convolutional neural networks with simple negative sampling[J]. arXiv preprint arXiv:1506.07650, 2015.

\bibitem{ref_Zelenko}
Zelenko, Dmitry, Chinatsu Aone, and Anthony Richardella. "Kernel methods for relation extraction." Journal of machine learning research 3.Feb (2003): 1083-1106.

\bibitem{ref_MiwaMakoto}
Miwa, Makoto, et al. "A rich feature vector for protein-protein interaction extraction from multiple corpora." Proceedings of the 2009 Conference on Empirical Methods in Natural Language Processing: Volume 1-Volume 1. Association for Computational Linguistics, 2009.

\bibitem{ref_Chan}
Chan, Yee Seng, and Dan Roth. "Exploiting syntactico-semantic structures for relation extraction." Proceedings of the 49th Annual Meeting of the Association for Computational Linguistics: Human Language Technologies-Volume 1. Association for Computational Linguistics, 2011.

\bibitem{ref_Yu}
Yu, Xiaofeng, and Wai Lam. "Jointly identifying entities and extracting relations in encyclopedia text via a graphical model approach." Proceedings of the 23rd International Conference on Computational Linguistics: Posters. Association for Computational Linguistics, 2010.

\bibitem{ref_Miwa}
Miwa, Makoto, and Yutaka Sasaki. "Modeling joint entity and relation extraction with table representation." Proceedings of the 2014 Conference on Empirical Methods in Natural Language Processing (EMNLP). 2014.

\bibitem{ref_Li}
Li, Qi, and Heng Ji. "Incremental joint extraction of entity mentions and relations." Proceedings of the 52nd Annual Meeting of the Association for Computational Linguistics (Volume 1: Long Papers). Vol. 1. 2014.

\bibitem{ref_Gupta}
Gupta, Pankaj, Hinrich Schutze, and Bernt Andrassy. "Table filling multi-task recurrent neural network for joint entity and relation extraction." Proceedings of COLING 2016, the 26th International Conference on Computational Linguistics: Technical Papers. 2016.

\bibitem{ref_Zheng}
Zheng, Suncong, et al. "Joint Extraction of Entities and Relations Based on a Novel Tagging Scheme." Proceedings of the 55th Annual Meeting of the Association for Computational Linguistics (Volume 1: Long Papers). 2017.

\bibitem{ref_Xiangrong}
Zeng, Xiangrong, et al. "Extracting relational facts by an end-to-end neural model with copy mechanism." Proceedings of the 56th Annual Meeting of the Association for Computational Linguistics (Volume 1: Long Papers). 2018.

\bibitem{ref_Xiaoya}
Li, Xiaoya, et al. "Entity-Relation Extraction as Multi-Turn Question Answering." arXiv preprint arXiv:1905.05529 (2019).

\bibitem{ref_Bekoulis}
Bekoulis, Giannis, et al. "Joint entity recognition and relation extraction as a multi-head selection problem." Expert Systems with Applications 114 (2018): 34-45.



\bibitem{ref_Lafferty}
Lafferty, John, Andrew McCallum, and Fernando CN Pereira. "Conditional random fields: Probabilistic models for segmenting and labeling sequence data." (2001).

\bibitem{ref_Makoto}
Miwa, Makoto, and Mohit Bansal. "End-to-end relation extraction using lstms on sequences and tree structures." arXiv preprint arXiv:1601.00770 (2016).

\bibitem{ref_Tianqi}
Chen, Tianqi, and Carlos Guestrin. "Xgboost: A scalable tree boosting system." Proceedings of the 22nd acm sigkdd international conference on knowledge discovery and data mining. ACM, 2016.

\bibitem{ref_symbert}
Cheng X, Xu W, Chen K, et al. Symmetric Regularization based BERT for Pair-wise Semantic Reasoning. arXiv preprint arXiv:1909.03405, 2019.

\end{thebibliography}
\end{document}